\documentclass[letterpaper, 10 pt, conference]{ieeeconf} 
\IEEEoverridecommandlockouts                              

\usepackage{graphicx}

\usepackage{amsmath} 
\usepackage{amssymb}  
\usepackage{subcaption}
\usepackage{paralist} 
\usepackage{xcolor}
\usepackage{url}
\usepackage{flushend} 
\usepackage{booktabs} 
\usepackage{color, colortbl} 
\definecolor{Gray}{gray}{0.9}

\title{\LARGE \bf
Neural Network Controller for Autonomous Pile Loading Revised
}

\author{Wenyan Yang$^{1}$, Nataliya Strokina$^{1}$, Nikolay Serbenyuk$^{2}$, Joni Pajarinen$^{3,4}$, \\ Reza Ghabcheloo$^{2}$, Juho Vihonen$^{5}$, Mohammad M. Aref$^{5}$ and Joni-Kristian K{\"a}m{\"a}r{\"a}inen$^{1}$ 
\thanks{$^{1}$Computing Sciences, Tampere University, Finland,
$^{2}$Automation Technology and Mechanical Engineering, Tampere University, $^{3}$Department of Electrical Engineering and Automation, Aalto University, $^{4}$ Intelligent Autonomous Systems, Technische Universität Darmstadt, Germany and $^{5}$ Cargotec Oyj}
}

\begin{document}
\maketitle
\thispagestyle{empty}
\pagestyle{empty}

\begin{abstract}



We have recently proposed two pile loading controllers that learn
from human demonstrations:
a neural network (NNet)~\cite{Halbach2019} and
a random forest (RF) controller~\cite{Yang-2020-icra}. In the field experiments 
the RF controller obtained clearly better success rates.
In this work, the previous findings are drastically revised by experimenting summer time trained controllers in winter conditions.
The winter experiments revealed a need for additional sensors, more training data, and a controller that can take advantage of these.
Therefore, we propose a revised neural
controller (NNetV2) which has a more expressive structure and uses a
neural attention mechanism to focus on important parts of the sensor and control signals.
Using the same data and sensors to train and test the three controllers,
NNetV2 achieves better robustness against drastically changing
conditions and superior success rate. To the best of our knowledge, this is the first work testing a learning-based controller for a heavy-duty machine in drastically varying outdoor conditions and delivering high success rate in winter, being trained in summer.
\end{abstract}

\section{INTRODUCTION}

\begin{figure}[t]
\centering
  \includegraphics[width=0.9\linewidth]{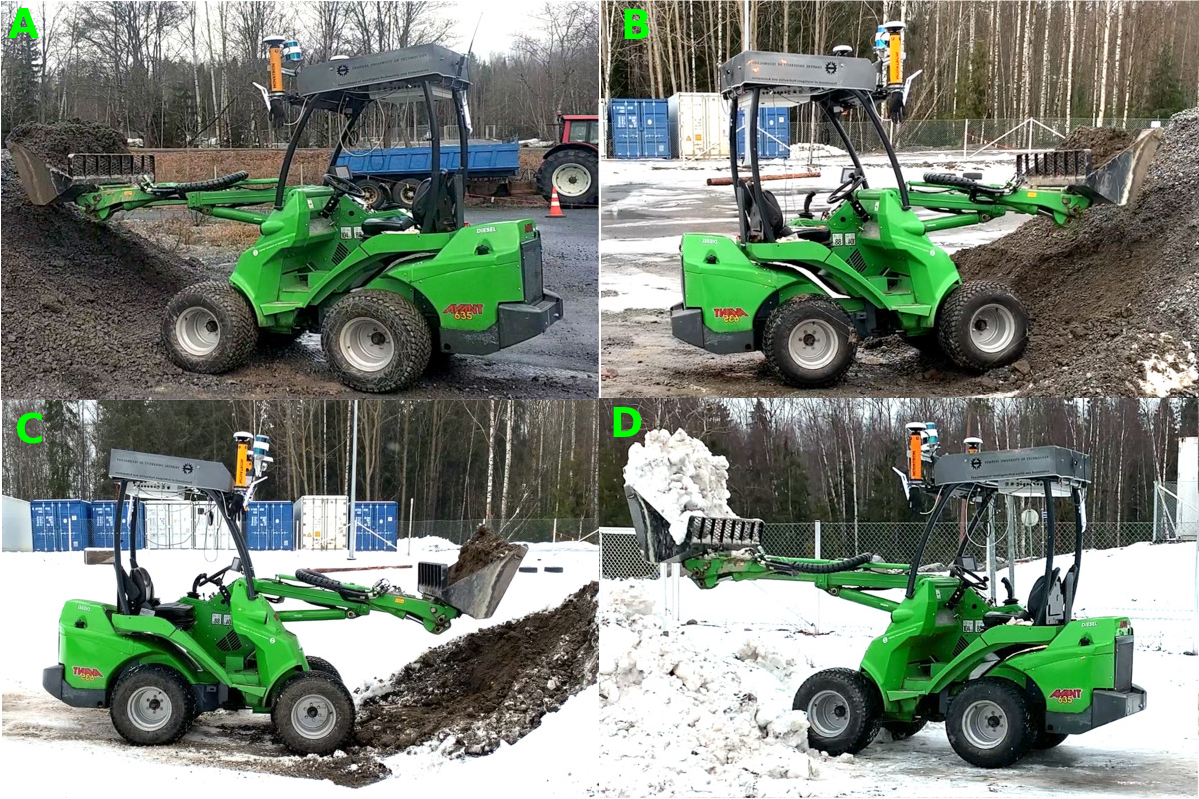}  
\caption{The training demonstrations were recorded during summer time
(previously used in~\cite{Yang-2020-icra}). The wheel loader test experiments were conducted during winter time in various conditions over 30 days: cold wet weather (A), mild icy surface and partial snow (B), icy surface and snow covering the surface and pile (C), cold winter conditions and gravel moving partly replaced with snow clearing (D).}
\label{fig:teaser}
\end{figure}

Pile loading is one of the most challenging tasks in earth moving automation for Heavy-duty mobile (HDM) machines. This is partly caused by  the difficulty of modelling the interaction between the tool and the material~\cite{Dadhich2016} and partly because of high variation in work sites and weather conditions throughout the year
(Fig.~\ref{fig:teaser}). Weather conditions affect the material properties, the hydraulics properties of the machine, and the ground surface properties. The state-of-the-art works on pile loading or excavation automation are either model-based or use heuristics~\cite{Sotiropoulos2020,Silver-2014-icml}, and experimented only
in simulators or with toy setups. Therefore it is unclear how well
these methods perform in real work sites.

Recently, there have been attempts to learn a
controller from human demonstrations using
machine learning techniques to approximate the controller.
Neural Network (NN) based controllers
are proposed in Dadlich~et~al.~\cite{Dadlich2019} and Halbach~et~al.~\cite{Halbach2019} and both were tested with
real wheel loaders.
Dadlich~et~al. concluded that {\em "different networks are needed to be trained with data collected in different conditions"}. That was
verified in the experiments of Yang~et~al.~\cite{Yang-2020-icra} who
showed that the NNet controller by Halbach~et~al. fails when
test conditions are changed (e.g. distance to the pile). As
a solution Yang~et~al. propose a Random Forest (RF) based controller that achieves clearly better success rate than
NNet.
Although RF performs well in the pile loading task, neural network
controllers have beneficial properties for machine learning in robotics.
In particular, neural networks are differentiable making them suitable
for autonomous learning and exploration using the popular
policy gradient Reinforcement Learning techniques~\cite{Sutton-2000-nips,Silver-2014-icml,lillicrap2015continuous}.


In this work, we propose neural network controllers (ANNet, DANNet) that are competitive against the prior arts.
The attention module selects important signals during the different states of the pile loading task.
In the winter experiments, these changes improve the performance of NNet  from 0\% to 100\%. In addition, our previous
results and findings 
are drastically revised thanks to the experiments conducted
in winter with controllers trained using summer demonstrations. 
Our work provides the following contributions:
%
\begin{compactitem}

\item A neural network controller with a neural attention module, which selects sensors that are important at different states of the pile loading control problem. It prevents neural controller's failure in changing environmental and dynamic load conditions.


\item 
 We highlight the loss of useful information in the cases of downsampling or "filtering" suitable input data (manual selection of the best demonstrations as done in the previous works), which have negative effect on the success rate.  



\item  
We demonstrate that the previous works lack controller state observability by experimentally comparing a variety of available sensors. Particularly in winter, the previously omitted hydraulic pressure at telescope joints plays a significant role in observability of load dynamics. 




\end{compactitem}
All controllers were implemented and experimented on a real-scale robotic wheel-loader. Testing was performed on multiple winter days over a 30 day period of time at different locations. Over this time the weather conditions changed dramatically, including an icy road, frozen material, wet snow and mud (see Fig.~\ref{fig:teaser}). This allowed us to verify the findings in highly diverse test conditions. The code and dataset will be published.

\section{RELATED WORK}

{\em Autonomous pile loading} works adopt heuristics~\cite{Fernando2018} or are model-based~\cite{Sotiropoulos2019}, and are 
experimented only in a simulator~\cite{Fernando2018} or toy-scale setups~\cite{Jud2017, Sotiropoulos2020}, which cannot capture the complicated phenomena of the real-world problem. 
 Jud et al.~\cite{Jud2017} utilize the trajectory of end-effector force-torque instead of the end-effector position to learn autonomous excavation. This way the model avoids generating arbitrarily high forces. In~\cite{Fernando2018} Fernando et al. present a heuristic algorithm to learn  an admittance controller for autonomous loading from a muck pile with non-homogeneous material. The proposed algorithm learns to apply specific forces instead of learning the desired trajectory.
 Sotiripoulos and Asada~\cite{Sotiropoulos2019} use the power transmitted from the excavator to the soil as an input for adaptive excavation algorithm.
 By maximizing the output product of force and velocity the method enables bucket filling control. In the follow-up work on a similar set-up, \cite{Sotiropoulos2020} presents a feedback controller for rock scooping that optimizes a cost function using a Gaussian Process model to predict rock behaviour.

Model-based approaches succeed in many robotics applications. However, in pile loading the interaction between the bucket and the material is hard to model accurately. Several works attempt to learn this interaction using {\em learning from demonstrations}. Dadhich et al.~\cite{Dadhich2016}  fit linear regression models to the lift and tilt bucket commands recorded with a joystick. Fukui et al.~\cite{Fukui2015} use a neural network 
model that selects a pre-programmed excavation motion from a dataset of motions.
\cite{Dadlich2019,Halbach2019,Yang-2020-icra} 
report real experiments of autonomous scooping with a real-scale HDM machine.
Dadlich et al.~\cite{Dadlich2019} propose a shallow time-delay neural network controller.
The controller uses the joint angles and velocities as inputs.
After outdoor experiments the authors conclude that for different
conditions the network controller needs to be retrained. Halbach et al.~\cite{Halbach2019} train a shallow neural network controller (NNet) for bucket loading based on the joint angles and hydraulic drive transmission 
pressure. Yang et al.~\cite{Yang-2020-icra}, presented a RF pile loading controller trained using a few demonstrations and the
same sensors as Halbach et al.
In field experiments RF clearly outperformed the NNet controller.
In this work we revise the findings of Halbach and Yang et al.
that are invalid if the training and test conditions are
drastically different.


\section{METHODS}\label{sec:methods}

In {\em learning from demonstrations} or {\em imitation learning} a controller is learned from human demonstrations. A number of sensor readings $\mathbf{s}_i$ are observed at each discrete time step $i$. For timestep $i$, The controller takes the sensor readings $\mathbf{s}_i$ as input and outputs control actions $\mathbf{u}_i$ that approximate the human
actions
The observation-action pairs constitute the training set $D$ of  demonstrations:
$D = \{\langle \vec{s}_i, \vec{u}_i\rangle\}_{i=1..T}$, where $T$ is the total number of samples. 
This yields a supervised learning problem, where the control actions are predicted by a function approximator $F$ with the parameters $\vec{\Theta}$:
%
\begin{equation}
\label{eq:controller}
\vec{u} = F(\vec{s};\vec{\Theta}) \enspace .
\end{equation}
The approximator function is optimized to fit to the expert demonstrations using a suitable loss function $\ell$:
%
\begin{equation}
\label{eq:loss}
\min_\mathbf{\Theta}{\sum_{i = 1}^{T}\ell\left(F(\mathbf{s}_i;\mathbf{\Theta}), \mathbf{u}_i\right)} \enspace .
\end{equation}
The standard loss is the Mean Squared Error (MSE)
\begin{equation}
\ell\left(F(\mathbf{s}_i;\mathbf{\Theta}), \mathbf{u}_i\right) = \frac{1}{T} \sum_{i=1}^{T}||F(\mathbf{s}_i;\mathbf{\Theta})-\vec{u}_i||_2^2 \enspace .
\end{equation}

\begin{figure*}[th]
    \centering
    \begin{subfigure}{0.18\textwidth}
      \centering
      \includegraphics[width=1\linewidth]{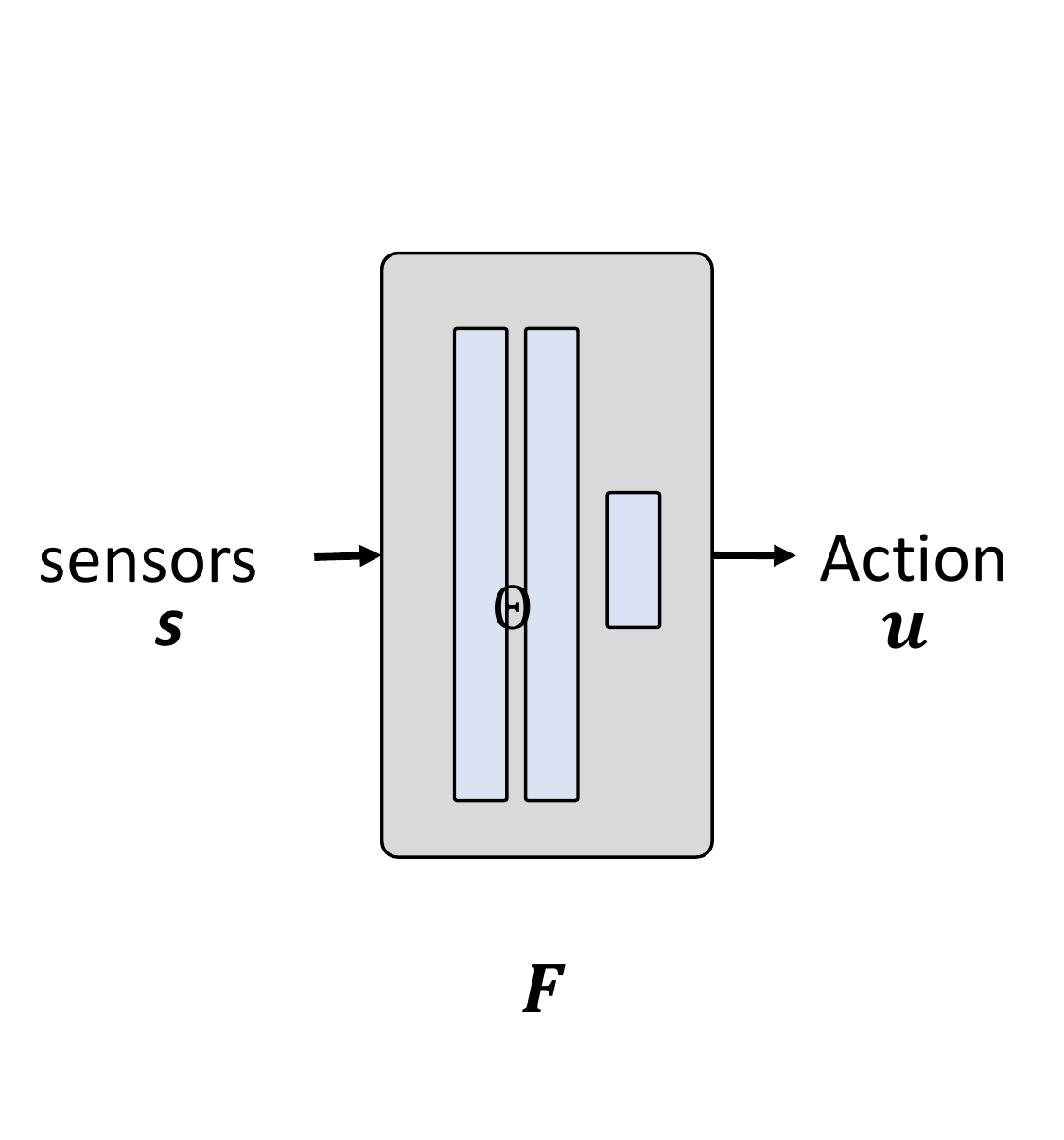}
      \caption{NNetV2}
    \end{subfigure}%
    \hfill
    \begin{subfigure}{0.3\textwidth}
      \centering
      \includegraphics[width=1\linewidth]{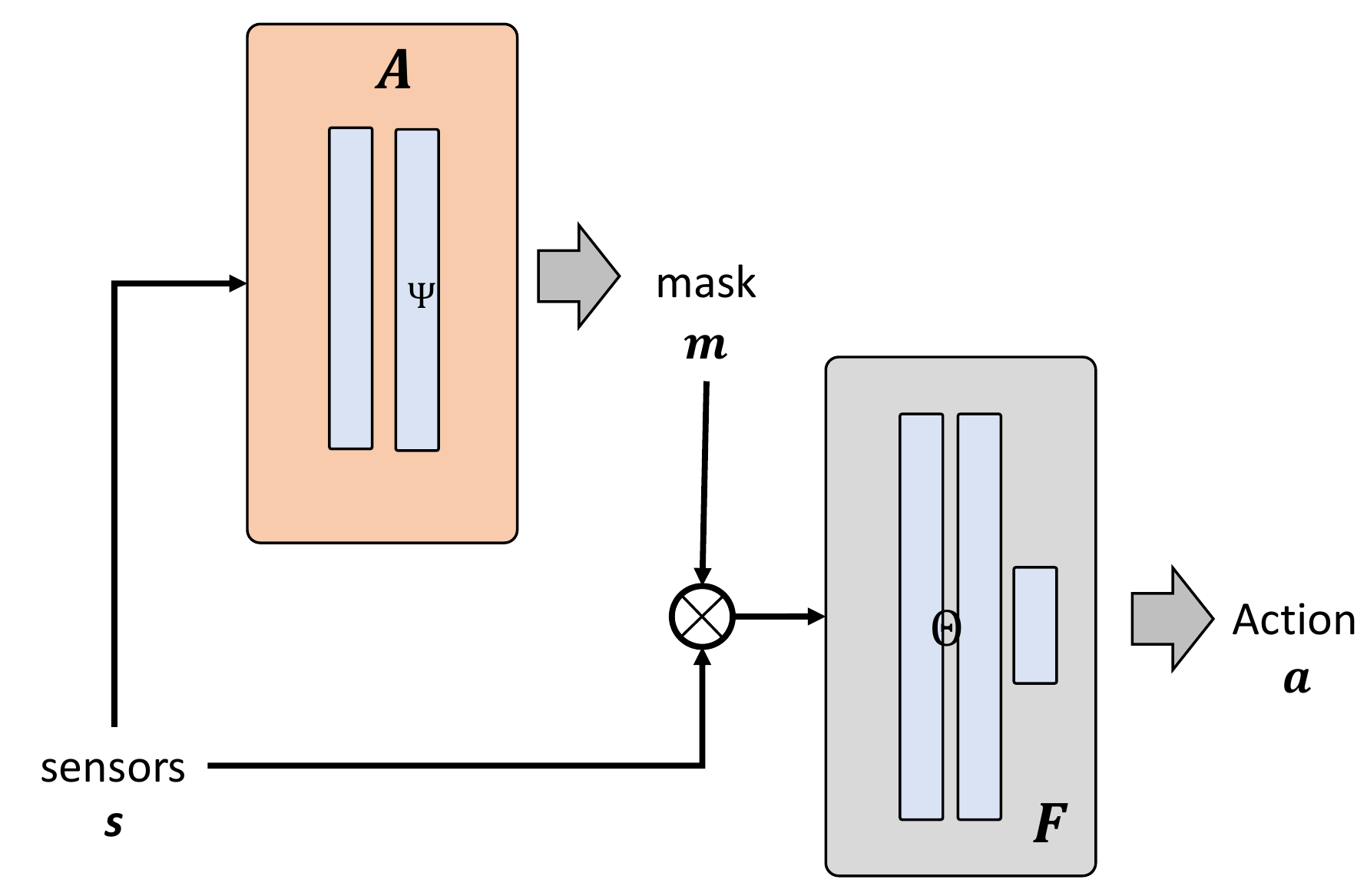}
      \caption{Single attention architecture (ANNet)}
    \end{subfigure}
    \hfill
    \begin{subfigure}{0.4\textwidth}
      \centering
      \includegraphics[width=1\linewidth]{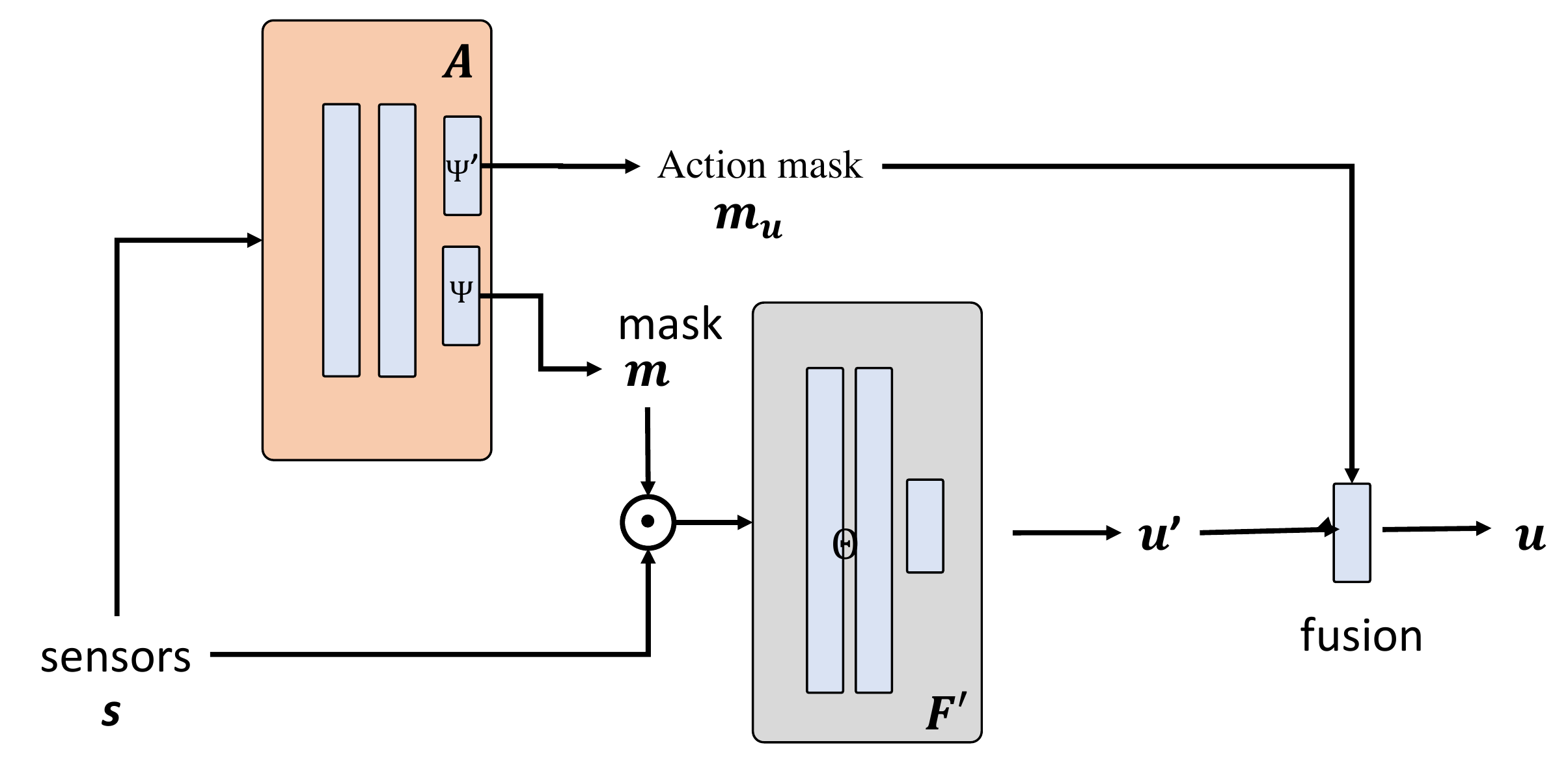}
      \caption{Dual attention (DANNet)}
    \end{subfigure}
    
  \caption {The three proposed neural network controller architectures experimented in this work.}
\label{fig:architectures}
\end{figure*}

\vspace{\medskipamount}\noindent\textit{NNetV2 -- } A popular choice for the approximator $F$ is a neural network.
Hallbach~et~al.~\cite{Halbach2019} propose a shallow
fully connected conventional Multi-layer Perceptron (MLP) regressor
 network~\cite{Bishop1995book}. Their network has only five neurons in a single
layer ($\vec{s}$-5-$\vec{u}$) and it is trained using the Levenberg-Marquardt (L-M) backpropagation. A small MLP trained  has very limited expression power to effectively represent complex control policies (verified in our simulations). 
We revised the NNet network to NNetV2 which has three orders of magnitude more weights (Fig.~\ref{fig:architectures}(a)). NNetV2
has 200 neurons on two full-connected layers ($\vec{s}$-200-200-10-$\vec{u}$). The number of the inputs $\vec{s}$ and outputs $\vec{u}$ are the same in NNet and NNetV2 and they correspond to the HDM sensor and control signals.
 
\vspace{\medskipamount}\noindent\textit{Neural attention -- } Neural attention
has many successful applications, for example, in computer vision~\cite{Mnih2014, Jetley2018},
natural language processing~\cite{Bahdanau2015}
and robotics~\cite{Rao2017,venkatesh2019one}. The main function
of the "attention module" is to strengthen features important
for the target task and suppress features that are less
important~\cite{Jetley2018}. For unseen test samples 
attention helps to attenuate noise produced by redundant sensors.

The purpose of applying attention module in this work was driven by following motivations: (1) attention module shall automatically select important features for corresponding actions, which improves robustness of the neural network controller against a changing environment and conditions; (2) the attention module is able to make the black box controller more explainable.


%
\subsection{Neural attention module design}\label{sec:attention}
The attention mechanism is implemented as a fully connected neural network structure ($\vec{s}$-64-64-$\vec{m}$) that takes the
sensor signals as input and applies an attention mask
$\vec{m}$ to the same
inputs before they are given to the NNetV2 controller (Fig.~\ref{fig:architectures}(b)). The attention
controller is denoted as "ANNet" in our experiments. As a
novel solution we
also experiment with "dual attention" (DANNet in Fig.~\ref{fig:architectures}(c)) that provides attention
masks for both the inputs and outputs.

\subsubsection{Attention Neural Network Controller (ANNet)}
\label{sec:annet}
The architecture of the proposed attention neural network controller is illustrated
in Fig.~\ref{fig:architectures}(b). It is a full-connected
network with ReLU and Dropout layers: $\vec{s}$-64-64-$\vec{m}$, where $\vec{s}$ and $\vec{m}$ refer to the input and mask vectors. 


The attention network $A$ takes the sensor signals
$\vec{s}$ as the input and outputs an "attention feature vector"
$\vec{f}$ as
%
\begin{equation}
\vec{f} = A(\vec{s}; \vec{\Psi}) \enspace ,
\end{equation}
where $\vec{\Psi}$ are the attention network parameters. In the
experiments, we investigate whether additional sensors improve
attention and in that case the input with additional sensors is
denoted as $\vec{s}'$.

The attention features are normalized by the softmax operation producing an
{\em attention mask} $\vec{m}$:
\begin{equation}
m_j  = softmax(f_j) 
= \frac{exp(f_j)}{\sum_{i=1}^{N}exp(f_i)}
\end{equation}
%
or more compactly written as:
\begin{equation}
\mathbf{m} = softmax\left(A(\vec{s}; \mathbf{\Psi})\right) \enspace .
\end{equation}

Finally, the original controller approximator equation
in (\ref{eq:controller}) is modified using dot product with the attention mask:
%
\begin{equation}
\mathbf{u} = F(\mathbf{s} \cdot \mathbf{m}; \mathbf{\Theta}) \enspace .
\end{equation}

The attention neural network controller ANNet is optimized using the following optimization problem
\begin{equation}
\min_{\vec{\Theta},\vec{\mathbf{\Psi}}}{\sum_{i=1}^{T}\ell\left(\vec{s}_i\cdot F(A(\vec{s}_i; \mathbf{\mathbf{\Psi}});\mathbf{\Theta}), \vec{u}_i\right)}
\enspace .
\end{equation}
ANNet is trained using the same MSE loss (Eq.~\ref{eq:loss}) as NNetV2.


%
%
\subsubsection{Dual Attention Neural Network Controller (DANNet)}
\label{sec:dannet}
%
\begin{figure}[h]
\centering
\includegraphics[width=.8\linewidth]{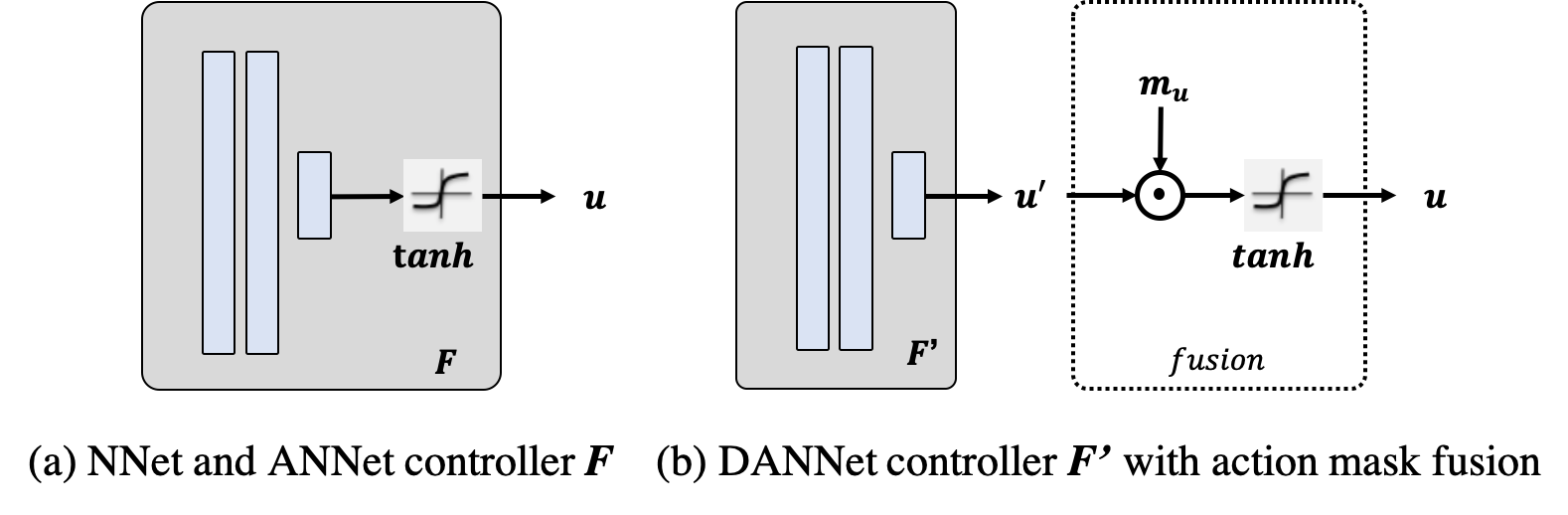}
\caption{The outputs of the both NNet and ANNet
are generated by the $\tanh$ nonlinearity (a). The
main difference is that in the DANNet
the output is first modulated (dot product) by the output attention vector $\vec{m}_u$ (b).}
\label{fig:dual_difference}
\end{figure}

For the DANNet architecture we introduce an additional attention mask $\vec{m}_u$ for the controller output. The control attention mask defines which control signals are "active". The intuition behind this idea comes from the demonstrations data itself - human
drivers rarely perform more than one action at the same time. In 88\% of the observation-action pairs in all recorded sequences there is only one action active.
The output action mask is generated from the same
sensor signals $\vec{s}$ as the input mask.
Therefore, the same attention network $A$ is
used, but its output is augmented to produce
an output attention feature vector of the size
of the control signal $\vec{u}$ as
$\vec{s}$-64-64-$\langle\vec{m},\vec{m}_u\rangle$
where
\begin{equation}
\mathbf{m}_{u} = softmax\left(A(\mathbf{s}; \mathbf{\Psi'})\right) \enspace ,
\end{equation}
$|\mathbf{m_u}| = |\mathbf{u}|$ and
$\vec{\Psi}'$ denotes the attention network
parameters trained for dual attention mechanism.

As a small difference for the DANNet as compared
to NNet and ANNet we add the control signal
mask $\vec{m}_u$ inside the last nonlinearity
function $\tanh$ as
%
\begin{equation}
\mathbf{u} =  \tanh(\mathbf{m_u} \cdot \mathbf{u'})
\end{equation}
where $\vec{u}'$ is the $F$ controller output
before the nonlinearity (Fig.~\ref{fig:dual_difference}). Otherwise,
the loss function and training procedures are
equivalent to NNetV2 and ANNet.


%

%
%
\subsection{Training details} \label{sec:nn_details}
The NNetV2 controller structure is -200-200-10- and the
attention module structure of ANNet and DANNet is -64-64-.
The
two last hidden layers are regulated by setting
their dropout probabilities to 0.35 and
ReLU is used after each hidden layer. The output layer
of the controller network $F$ has three units with
$\tanh$-activation and it thus produces output control
signal vector $\mathbf{u} \in [-1.0, +1.0]_{3\times 1}$ that corresponds
to normalized delta velocities (increment/decrease)
of the control variables.

The attention module $A$ uses the same sensor
signals $\vec{s}$ except in the extra experiments where
the attention module is augmented with additional
sensors (see Table~\ref{tab:signals}).
%
Output control (action) vector $\vec{u}$ for all models is a three-dimensional vector: $\vec{u} = \langle{u_{\theta_1}, u_{\theta_2}, u_g}\rangle$ that denotes the delta control variables (velocities). All models were trained using the RAdam (Rectified Adam) optimizer~\cite{liu2019variance} with the mini-batch size 512 and the initial learning rate set to 0.001. The network weights were initialized using the Kaiming initialization~\cite{he2015delving}. All training converged after 150 epochs (see Section~\ref{sec:offline_validation} for more details).  

\begin{figure}[bth]
\centering
\includegraphics[width=0.9\linewidth]{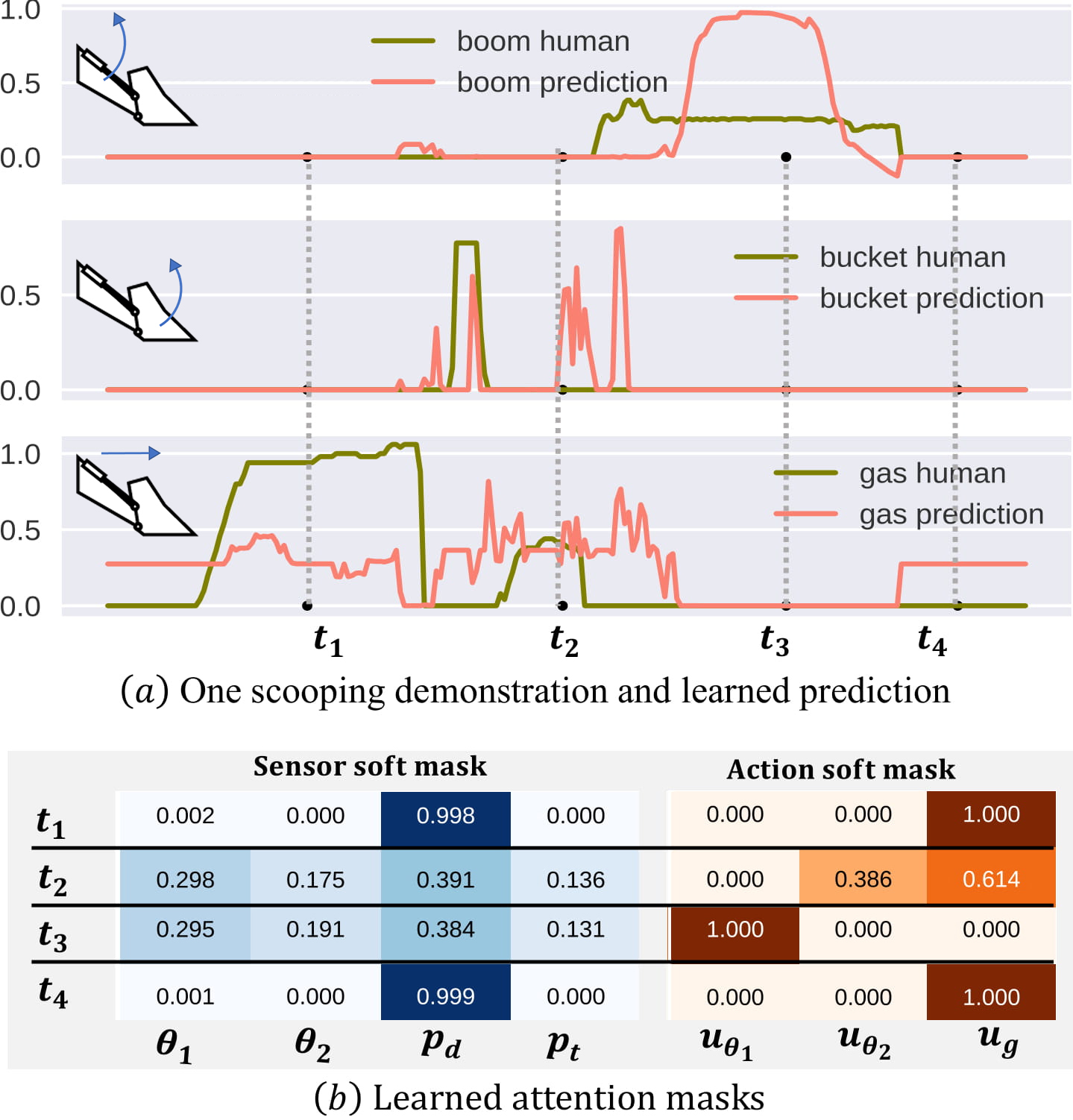}
\caption{Validation of the input (sensor) and output (control) masks. (a) human and DANNet control signals during a single pile loading action and (b) input and output attention masks at the four time stamps:
$t_1$ (drive toward the pile), 
$t_2$ (impact with the pile),
$t_3$ (bucket filling with boom) and
$t_4$ (finished).}
\label{fig:attention_masks}
\end{figure}

Fig.~\ref{fig:attention_masks}(a) demonstrates pile loading action control signals for human and DANNet and Fig.~\ref{fig:attention_masks}(b) shows the attention values for different control signals
at the time stamps $t_1$-$t_4$. At $t_1$, the main sensor attention is on the driving pressure $p_d$ and action attention on the gas command $u_g$ (approach pile). At time stamp $t_2$ and $t_3$, the sensor attention has moved to all input sensors while the action attention prefers the bucket $u_{\theta_2}$ and boom rising $u_{\theta_1}$ actions (raise the boom).


%
%





\section{EXPERIMENTS}\label{sec:experiments}


\vspace{\medskipamount}\noindent\textit{Overview of the experiments -- } The robot hardware setup is described in Sec.~\ref{sec:hardware_setup}. The experimental setup is introduced in Sec.~\ref{sec:experiment_setup}. The
first experiment (Experiment~1) compares the three controllers trained
with the original summer data and tested in winter conditions in Sec.~\ref{sec:experiments1}. In Experiment~2 the winter experiments are
repeated using the full original dataset (no down-scaling and
no manual selection) and with the additional sensors to improve
observability (Sec.~\ref{sec:experiments2}). Further ablation
study on the neural attention modules is presented in
Sec.~\ref{sec:experiments3} with real data and more detailed analysis
with offline validation data in Sec.~\ref{sec:offline_validation}.

%
\subsection{Robot hardware}  \label{sec:hardware_setup}
\vspace{\medskipamount}\noindent\textit{Wheel loader -- }
The scooping (bucket filling) experiments were conducted on Avant 635, a commercial wheel loader with articulated-frame steering, except its hydrostatic transmission power with by-wire controllers. It carries a three degrees-of-freedom manipulator in the vertical plan consisting of a boom, telescope, and bucket joints collinear with each other. Its bucket is positioned in vertical plane by two joints, the boom joint and the bucket joint, and in horizontal plane by drive (throttle/gas) and articulated by a frame steering mechanism.

\vspace{\medskipamount}\noindent\textit{Sensors -- }
The sensor and control signals used in this work are summarized
in Table~\ref{tab:signals}.
The main difference to the previous works~\cite{Halbach2019,Yang-2020-icra} is that there
is an additional pressure sensor, {\em hydraulic pressure at the telescope joint}, which importance
for robustness we experimentally demonstrate.

%
\begin{table}[h]
\caption{Avant wheel loader sensor and control signals (the necessary sensor for winter conditions is highlighted).}
\label{tab:signals}
\begin{center}
\resizebox{0.8\linewidth}{!}{
 \begin{tabular}{ll} 
    \toprule    
    \multicolumn{2}{l}{\em Sensor signals $\vec{s}$}\\
    $\theta_1$ & Boom joint angle \\
    $\theta_2$ & Bucket joint angle \\
    $p_d$ & Hydraulic drive transmission pressure\\
    \rowcolor{lightgray}
    $p_t$ & Hydraulic pressure at the telescope joint\\
    \midrule
    \multicolumn{2}{l}{\em Additional attention signals $\vec{s}'$}\\
    $p_l$ & Hydraulic pressure at the boom joint\\    
    $p_b$ & Hydraulic pressure at the bucket joint\\
    $a$   & HST pump angle, proportional to driving speed\\
    \midrule
    \multicolumn{2}{l}{\em Control signals $\vec{u}$}\\  
    $u_{\theta_1}$   & Boom joint control\\
    $u_{\theta_2}$  & Bucket joint control \\
    $u_g$ & Throttle (gas) command \\

 \bottomrule
\end{tabular}}
\end{center}
\end{table}
%


\vspace{\medskipamount}\noindent\textit{Controller hardware -- } 
The control system is composed of multiple layers. In the actuator level and on control area network (CAN), industrial micro-controllers implement the power management and safety functions. In the PC control level, a Simulink Realtime target executes time-critical modules such as localization. Sub-systems communicate sensor data and control commands via UDP running on a Jetson AGX Xavier (8-Core ARM v8.2 64-bit NVIDIA Carmel CPU and 512-core NVIDIA Volta GPU with 64 Tensor Cores) on-board. The data collection, 
and closed-loop control are implemented on Jetson. Learning is performed offline on a standalone machine. Sensor data was received by a separate UDP thread at 20Hz rate. Overall system performance was about 8Hz, but it was reduced to 3Hz to make execution of commands more feasible for Avant. 


%
\subsection{Experimental setup} \label{sec:experiment_setup}

\vspace{\medskipamount}\noindent\textit{Bucket filling task -- }
The experiments were conducted at an outdoor test site. The human demonstrations (training data) were the same as in our previous work~\cite{Yang-2020-icra} and therefore all results are comparable. All test experiments were conducted in the period of one month in winter conditions 
(Fig.~\ref{fig:teaser}). During the experiments the test site ground  was frozen, muddy, slippery or dry varying on
each day. The material properties changed as well, for example, partially frozen gravel, moist gravel and wet snow. The wheel loader performed the task learnt by the controller - drive up to the pile (varying distance and angle) and perform a scoop.

\vspace{\medskipamount}\noindent\textit{Performance measure -- }
The bucket load after each test run was manually classified to be either successful scoop or not. A successful scoop was recorded when the bucket was at least half-full and otherwise a failure was marked.
For all experiments, we report the success rate,
$\frac{N_{full}}{N}\cdot 100\%$
as the performance indicator. The test runs were conducted on multiple different days over a period of 30 days in different weather conditions. The distance to the pile was varied between one to five meters and the wheel loader was positioned approximately toward the pile. 

\vspace{\medskipamount}\noindent\textit{Training data -- }
Training data consists of the 72 demonstrations from
\cite{Yang-2020-icra} collected during the summer of 2019, where
52 of the demonstration finish with full bucket and are
therefore "ideal demonstrations".
The low-level sensor measurements were down-sampled to 
synchronize them with the video input (20~Hz). Using the same
data we define two different training sets:

\begin{compactitem}
    \item $\text{D}_\text{I}$: All 72 recorded human demonstrations using the original 500Hz sampling
    frequency for all sensor signals
    (total of 709,368 samples).
    
    \item $\text{D}_\text{II}$: Manually selected and temporally down-sampled (20Hz) data based on $\text{D}_\text{I}$ (52 best demonstrations all finishing the bucket full).  Total of 16,322 samples (observation-action pairs) are available.
  
\end{compactitem}

\vspace{\medskipamount}\noindent\textit{Controller structures -- } NNet~\cite{Halbach2019} has one hidden layer with 5 neurons. NNetV2 is described in Sec.~\ref{sec:methods}. RF controller~\cite{Yang-2020-icra} has 20 random trees and the maximum depth is 30. 

\subsection{Experiment 1: Transfer from summer to winter}\label{sec:experiments1}

As can be seen from Table~\ref{tab:summer_vs_winter}, both NNet, NNetV2 and RF controllers failed in the winter experiments using the data collected in summer for the training. The main failure cases were early boom rising and no boom rising at all.The results indicate that with such a big change in conditions the controllers trained by supervised learning simply cannot generalize. The results indicate both simple MLP-like neural network and RF controllers cannot be generalized.
 
 
\begin{table}[th]
\caption{Success rates of the controllers.
Winter experiments consist of 15 test runs for each controller and
are conducted over the time period of one month. Inputs of the controllers are $<\theta_1, \theta_2, p_d>$. }
\label{tab:summer_vs_winter}
\begin{center}
\resizebox{0.8\linewidth}{!}{
 \begin{tabular}{l r r r} 
    \toprule
    & \multicolumn{3}{c}{{\bf Train}: summer dataset $\text{D}_\text{II}$}\\
    & NNet~\cite{Halbach2019} & RF~\cite{Yang-2020-icra} & NNetV2\\
    \midrule
    {\bf Test}: summer & 0\% & 80\% & 12\% \\
    {\bf Test}: winter &  0\% &  40\% & 0\%\\
    \bottomrule
\end{tabular}}
\end{center}
\end{table}


%
%
\subsection{Experiment 2: Adding data and telescopic joint pressure} \label{sec:experiments2}

In this experiment, we consider the dataset $\text{D}_\text{II} $ as the baseline dataset. We first added more data simply by using all recorded demonstrations to create a new dataset $\text{D}_\text{I} $:
\begin{compactitem}
    \item including the ones with the half full bucket
    \item keeping the original data recorded at the rate of
500Hz.
\end{compactitem}
Except expanding the amount of training pairs, we also added one extra input sensor dimension $p_t$ to the input sensors $\vec{s}=\langle\theta_1, \theta_2, p_d, p_t\rangle$ used in the previous work
$\vec{s}=\langle\theta_1, \theta_2, p_d\rangle$.

\begin{table}[th]
\caption{Success rates using all data and with and
w/o the additional pressure sensor $p_t$
(hydraulic telescopic pressure). Approximately
30 attempts were executed with each controller over the period of 30 days.} 
\label{tab:data_and_sensors}
\begin{center}
\resizebox{0.7\linewidth}{!}{
 \begin{tabular}{l l r r} 
    \toprule    
    {\em Controller} & $p_t$ & \multicolumn{2}{c}{Training data (summer)}\\
                    & & {$\text{D}_\text{II} $~\cite{Yang-2020-icra}}
                    & {$\text{D}_\text{I} $} \\
    \midrule
    RF              &  & 40\% &    30\%\\
    RF              & \checkmark &   86\% &    87\%\\
    NNet            &  & 0\% &    0\%\\
    NNet            & \checkmark & 0\% &    0\%\\
    NNetV2          &  & 0\% &    56\% \\
    NNetV2          &  \checkmark & 0\% &    76\% \\
 \bottomrule
\end{tabular}}
\end{center}
\end{table}

\vspace{\medskipamount}\noindent\textit{Effect of expanded dataset $\text{D}_\text{I} $ -- }
The bigger neural network NNetV2 controller obtained a clear improvement from 0\% to 56\% with $\text{D}_\text{I} $. The simple NNet\cite{Halbach2019} was not either improved by additinal sensor dimension nor the expanded dataset.

\vspace{\medskipamount}\noindent\textit{Effect of additional sensor $p_t$ -- } With the additional sensor, $p_t$, NNetV2 was improved from 56\% to 76\% and RF from 30\% to 87\%. Clearly the hydraulic telescopic pressure sensor provides important information that makes the unknown controller states more observable. Our finding is that the original drive transmission pressure $p_d$ is affected by wheel slip on icy surface while the telescopic joint pressure remains unaffected and correctly triggers boom rise.

\subsection{Experiment 3:  Neural attention} \label{sec:experiments3}

\begin{table}[h]
\caption{Success rates of the three proposed neural network controllers: NNetV2, single attention network ANNet and double attention network DANNet (see Section~\ref{sec:attention}).
$\vec{s}'$ denotes usage of the additional attention sensors (Table~\ref{tab:signals}).} 
\label{tab:attention}
\begin{center}
\resizebox{0.8\linewidth}{!}{
 \begin{tabular}{l l l r r} 
    \toprule    
    {\em Controller} & \multicolumn{4}{c}{Training data (summer)}\\
                    & $p_t$ & $\vec{s}'$ & {$\text{D}_\text{I} $} & {$\text{D}_\text{II} $}\\
    \midrule
    NNetV2          &  & & 56\% &    0\% \\
    ANNet           &  & & 48\% &    0\% \\
    DANNet          &  & & 24\% &    0\% \\
    \midrule
    NNetV2     & \checkmark & & 76\% &    0\% \\
    ANNet      & \checkmark & & 100\% &   60\%\\
    DANNet     & \checkmark & &  100\% &  100\%\\
    \midrule
    NNetV2      & \checkmark & \checkmark     & 0\% &  0\% \\
    ANNet      & \checkmark & \checkmark     & 100\% &  80\% \\
    DANNet     & \checkmark & \checkmark    & 100\% & 100\%\\
 \bottomrule
\end{tabular}}
\end{center}
\end{table}
In Section~\ref{sec:attention} we introduce ANNet
that learns to mask the input signals with the attention mask and DANNet that masks both the input sensor and output control signals.
The results for all proposed neural network controllers
are shown in Table~\ref{tab:attention}. There are two important
findings:  with $p_t$ given, attention module boost neural controller's performance even trained with $\text{D}_\text{II} $; with $s'$ given to attention module, both ANNet and DANNet reach $100\%$ success rate. Comparing with RF in Table~\ref{tab:data_and_sensors}, the ANNet and DANNet provides on par or even
better performance.

\subsection{Experiment 4: Simulation studies}
\label{sec:offline_validation}
\begin{figure}[h]
    
    \centering
    \begin{subfigure}{0.8\linewidth}
      \centering
      \includegraphics[width=1\linewidth]{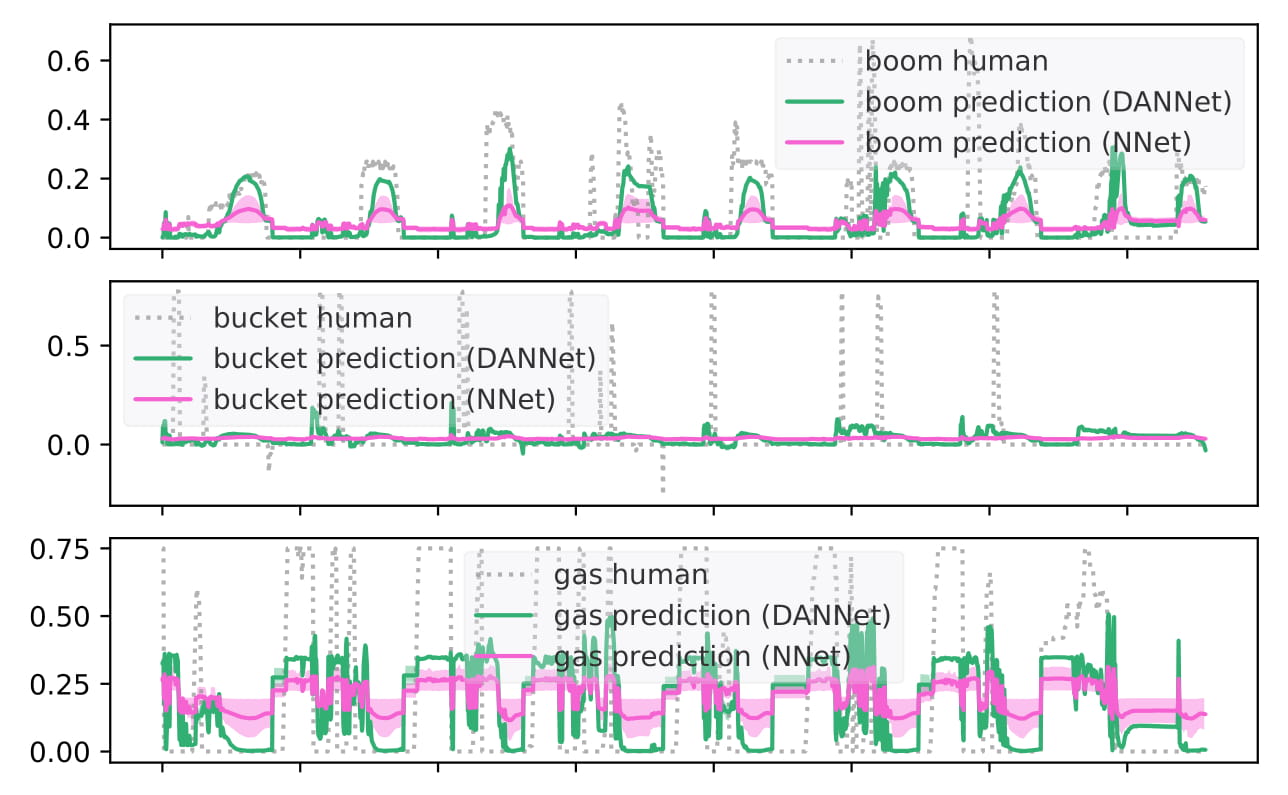}
    \end{subfigure}
  \caption {Comparison of the output control signals for the proposed neural network controllers NNetV2, DANNet and the ground truth (human demonstration).\textcolor{blue}{Additional figures are provided within the supplementary video}} 
\label{fig:test_conditions}
\end{figure}

\begin{figure}[h]
    \centering
    \begin{subfigure}{0.8\linewidth}
      \centering
      \includegraphics[width=1\linewidth]{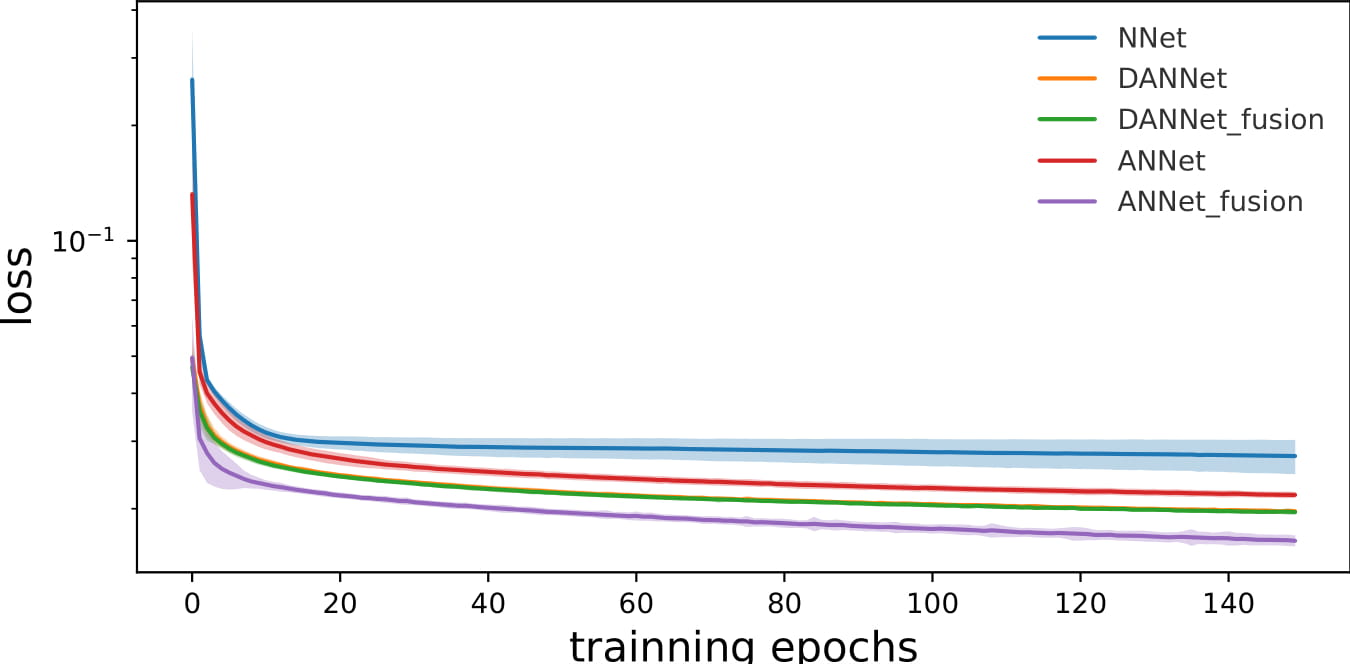}
      \caption{Trained with the ideal dataset $\text{D}_\text{II} $ in
      \cite{Yang-2020-icra}}
    \end{subfigure}
    
    \vspace{\medskipamount}
    
    \begin{subfigure}{1\linewidth}
      \centering
      \includegraphics[width=0.8\linewidth]{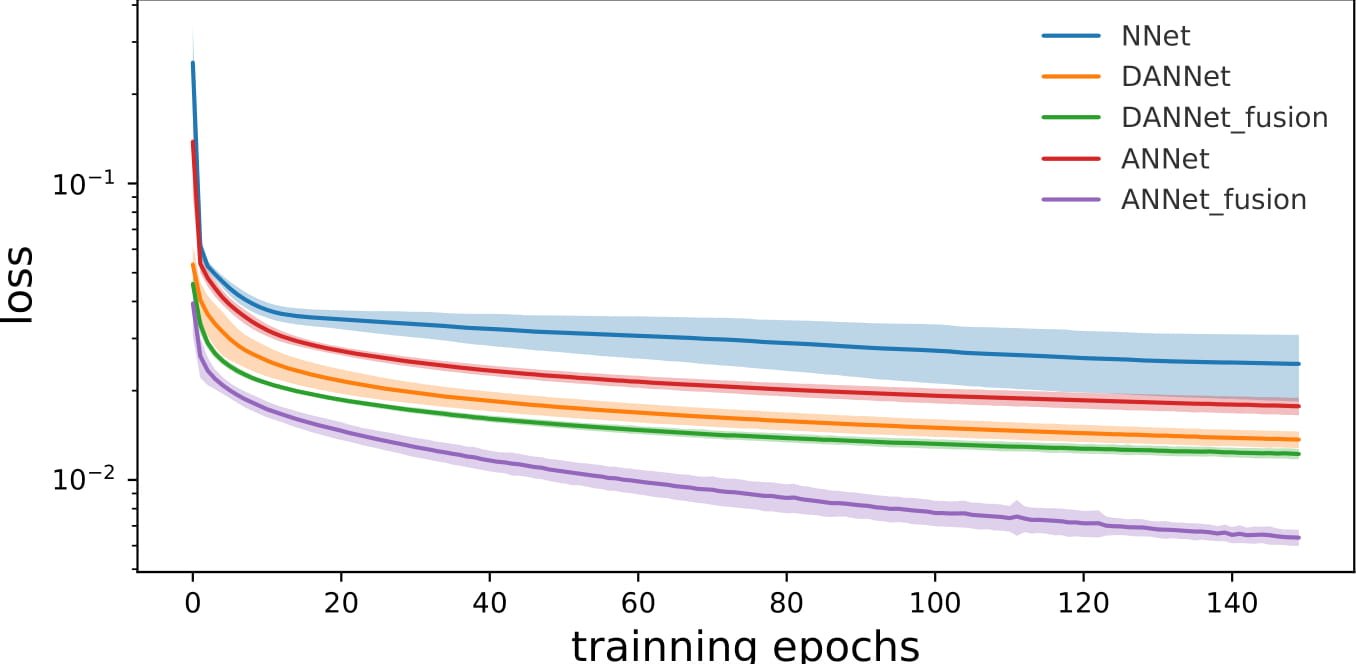}
      \caption{Trained with all data $\text{D}_\text{I} $}
    \end{subfigure}
  \caption {The mean (solid line) and standard deviation (shading) test
  data error (validation loss) values for the proposed neural controllers.
    Values are computed over 100 trials using the same summer training data, random initialization
  and the same offline winter test data (10 human demonstrations). ANNet\_fusion and DANNet\_fusion refers to training with $s'$}
\label{fig:training_loss}
\end{figure}

10 recorded human demonstrations
in winter conditions allow us to study behavior of the controllers more analytically. Fig.~\ref{fig:training_loss} shows the validation loss during training.
Two important findings that verify the results with the real pile loader: 1) the MSE loss between the predicted and ground truth control signals obtains smaller test set error with all available training data (Fig.~\ref{fig:training_loss}(a) and (b));
2) the attention networks, ANNet and DANNet, always obtain lower MSE than the networks without attention and the attention networks' converge is more stable.
These findings can be qualitatively verified in Fig.~\ref{fig:test_conditions} where
the both ANNet and DANNet produce control signals that match
the human demonstrations better than NNetV2.

%
\section{CONCLUSION}\label{sec:conclusion}

This work presents new results and findings for learning a pile loading controller
from human demonstrations. The previously proposed
neural network controller NNet~\cite{Halbach2019} and random forest controller
RF~\cite{Yang-2020-icra} fail if the test conditions are drastically
different from the training conditions (see Experiment~1). The failures of
the neural controller can be fixed by using a more expressive architecture (NNetV2) and by adopting modern deep learning optimization and non-linearities
(Section~\ref{sec:methods}). In addition, NNetV2 benefits from more data and
from a sensor that makes the control problem more observable (Experiment~2). Finally,
the neural network controllers adapting the attention mechanism, ANNet and DANNet, achieve superior results as compared to NNetV2 (Experiment~3).
The attention network controllers
produce signals that match more accurately to human behavior
(Experiment~4). Overall, we are convinced that the proposed attention
network controllers ANNet and DANNet are suitable for the task
of learning pile loading from demonstrations. Our future work will focus on
performing high level tasks that are autonomously learned using
reinforcement learning on top of the low level controllers learned by demonstrations.

\bibliographystyle{ieeetr}
\bibliography{root}


\end{document}